\documentclass[runningheads]{llncs}
\usepackage{graphicx}
\usepackage{todonotes}
\usepackage{subcaption}
\usepackage{subcaption}
\usepackage{hyperref}       % hyperlinks
\usepackage{url}            % simple URL typesetting
\usepackage{booktabs}       % professional-quality tables
\usepackage{amsfonts}       % blackboard math symbols
\usepackage{nicefrac}       % compact symbols for 1/2, etc.
\usepackage{microtype}      % microtypography

\begin{document}
\title{Task Fingerprinting for Meta Learning \\in Biomedical Image Analysis}

\author{Patrick Godau\inst{1,2}\orcidID{0000-0002-0365-7265} \and
Lena Maier-Hein\inst{1,2}\orcidID{0000-0003-4910-9368}}
% index{Godau, Patrick}
% index{Maier-Hein, Lena}
\authorrunning{P. Godau and L. Maier-Hein}
\institute{German Cancer Research Center, Heidelberg, Germany 
\email{patrick.scholz@dkfz-heidelberg.de} \and
Ruprecht-Karls University  of Heidelberg, Heidelberg, Germany}
\maketitle              % typeset the header of the contribution
\begin{abstract}
Shortage of annotated data is one of the greatest bottlenecks in biomedical image analysis. Meta learning studies how learning systems can increase in efficiency through experience and could thus evolve as an important concept to overcome data sparsity. However, the core capability of meta learning-based approaches is the identification of similar previous tasks given a new task - a challenge largely unexplored in the biomedical imaging domain. In this paper, we address the problem of quantifying task similarity with  a concept that we refer to as \textit{task fingerprinting}. The concept involves converting a given task, represented by imaging data and corresponding labels, to a fixed-length vector representation. 
In fingerprint space, different tasks can be directly compared irrespective of their data set sizes, types of labels or specific resolutions. 
An initial feasibility study in the field of surgical data science (SDS) with 26 classification tasks from various medical and non-medical domains suggests that task fingerprinting could be leveraged for both (1) selecting appropriate data sets for pretraining and (2) selecting appropriate architectures for a new task.  Task fingerprinting could thus become an important tool for meta learning in SDS and other fields of biomedical image analysis.

\keywords{Meta learning \and Knowledge transfer \and Medical image analysis \and Surgical data science \and Task similarity}
\end{abstract}
\section{Introduction}\label{sec_intro}
Shortage of annotated data is one of the greatest bottlenecks related to biomedical image analysis in general, and surgical data science (SDS) in particular \cite{maier-hein_sds_2020}. Methods proposed to address this issue include transfer learning  \cite{Cheplygina2019NotsosupervisedAS,Raghu2019TransfusionUT}, crowdsourcing  \cite{MaierHein2014CanMO} and self-supervised learning \cite{Ro2018ExploitingTP}. More recently, first attempts to leverage the concept of meta learning have been made \cite{Yuan2020FewIE,Zhang2020GeneralizingDL}. Meta learning studies how learning systems can increase in efficiency through experience \cite{Vilalta2005APV}, where experience can be represented by solutions to tasks connected to previously acquired data, for example. A core capability of meta learning-based approaches is the identification of similar previous tasks given a new task \cite{Hospedales2020MetaLearningIN}. Previous work has pioneered the idea of quantifying similarity between tasks in the domain of biomedical image analysis \cite{Cheplygina2017ExploringTS}. However, their proposed distance between any two given tasks depends on the set of tasks currently available. In consequence, re-computation of all distances is required every time a new task is added, rendering the approach unscalable. Further, validation was performed for a relatively small set of tasks. 

\begin{figure}
  \centering
\includegraphics[width=0.75\textwidth]{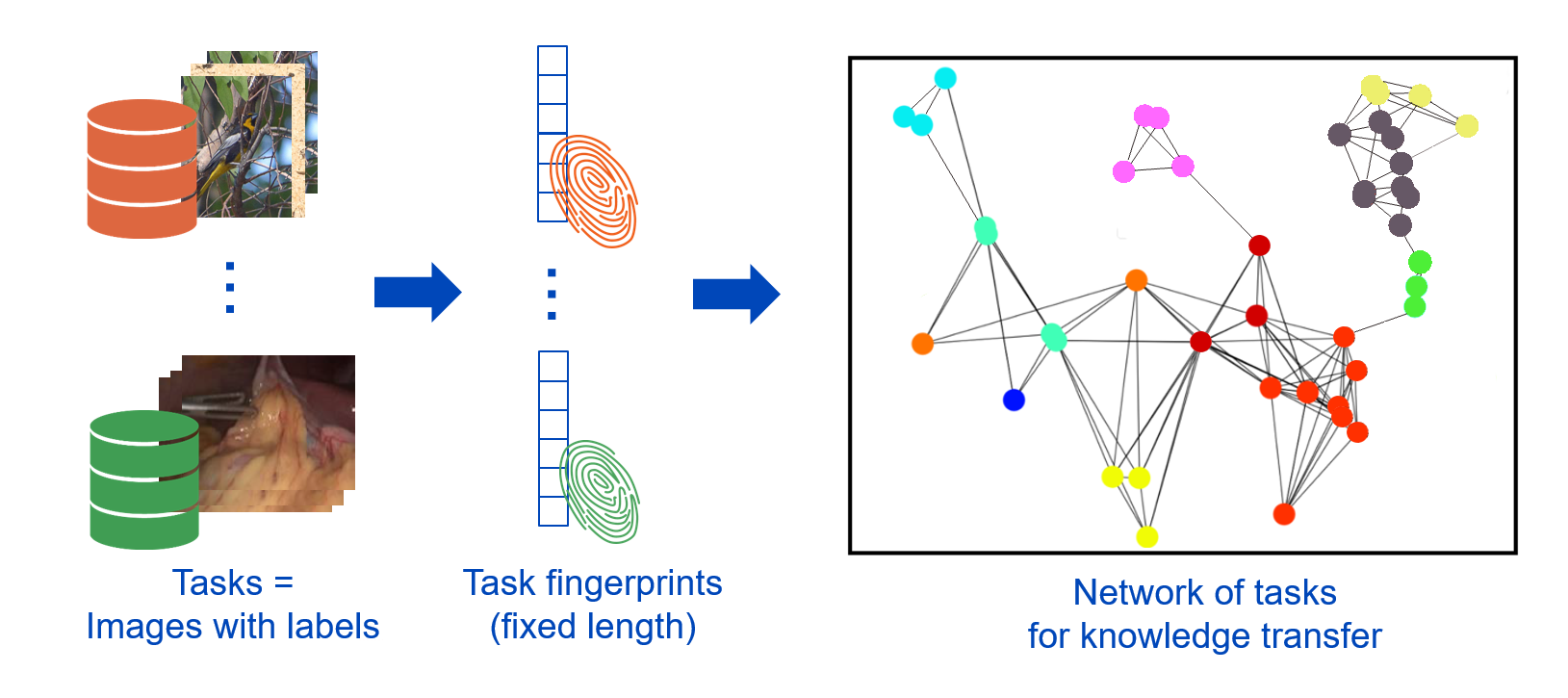}
  \caption{Concept of biomedical \textit{task fingerprinting}. A given task, represented by imaging data and labels, is converted to a fixed-length vector representation such that similarity of fingerprints indicates potential knowledge transfer.}
  \label{fig_concept}
\end{figure}

We address this gap in the literature from both a methodological and a validation perspective. As illustrated in Fig.~\ref{fig_concept}, we present the concept of biomedical \textit{task fingerprinting}, which involves representing a task (comprising images and labels), by a vector of fixed length irrespective of data set size, types of labels or specific resolutions. For the task embedding, we investigate several complementary approaches, which can be assigned to one of the two following categories: (1) those leveraging images \textit{and labels} for the embedding and (2) those directly comparing the distributions of the images with sample-based and optimal transport-based methods. Although all following methods could be transferred to a wide range of task types, we restrict ourselves to visual classification tasks as the primary scope. Visual classification tasks are relevant for a multitude of medical applications, such as the classification of colonoscopic or dermoscopic data for cancer screening or the classification of medical instruments for action recognition during surgery.

Our validation was performed on a total of 26 tasks from various domains with a focus on endoscopic vision (laparoscopy, colonoscopy, laryngoscopy). Specifically, we investigated the following hypotheses in the context of medical image analysis:
\begin{description}
\item[H1:] Task fingerprinting can capture semantic relationships between tasks. 
\item[H2:] Task fingerprinting can be used to select appropriate pretraining tasks for a given new task.
\item[H3:] Task fingerprinting can be used for model selection for a given new task. 
\end{description}

\section{Methods}\label{sec_methods}
This section presents the proposed concept of task fingerprinting along with the experimental conditions.

\subsection{Task fingerprinting}
We represent a \textbf{task} by $\mathfrak{T} = \{(x_i, y_i)\}_{i=1}^{N}$ comprising $N$ samples, each consisting of an image $x_i$ and a label $y_i$, as proposed by Achille et al. \cite{achille2019information}. We assume that the images corresponding to different tasks are of the same dimension after a homogenization process, detailed in Sec. \ref{sec_data}. Our aim is to find a corresponding embedding $\mathbf{e}_\mathfrak{T}$ together with a computable distance function $d$, such that for any two tasks $\mathfrak{T}, \mathfrak{T}'$: $d(\mathfrak{T}, \mathfrak{T}') :=  d(\mathbf{e}_{\mathfrak{T}}, \mathbf{e}_{\mathfrak{T}'})\geq 0$ and $d(\mathfrak{T}, \mathfrak{T}')$ is an indicator of \textit{how related} the two tasks are (defined in Sec. \ref{sec_design}). For defining $\mathbf{e}_\mathfrak{T}$ and $d$ we investigate a total of four complementary approaches: an embedding-based approach based on the Fisher Information Matrix (FIM) that leverages both images and labels and three approaches based on comparing only the image distributions:

\subsubsection{Fisher Embedding Distance (FED)} Inspired by Achille et al. \cite{achille2019task2vec},
we generate a task embedding that encodes both,  data and  labels, in one joint feature vector of fixed length by making use of the FIM. More precisely, for each task $\mathfrak{T}$, we use a pretrained probe-network (here: a standard ResNet34 \cite{He2016DeepRL}, that has been trained on ImageNet \cite{Deng2009ImageNetAL}) and retrain the final layer for the current task. For this tuned model, we then calculate the diagonal elements of the FIM, which is defined as: 
\begin{equation}
    F := \mathbb{E}_{x, y \sim \hat{p}(x), p_w(y | x)} [\nabla_w \log p_w(y | x) \nabla_w \log p_w(y | x)^T]
\end{equation}
with $\hat{p}$ being the empirical distribution of images in the task and $p_w$ being the prediction distribution of the trained model with respect to the weights $w$.
The resulting vector $\mathbf{e}$, with $\mathbf{e}_i := F_{i, i}$, represents the embedding of $\mathfrak{T}$. 
Note that we simplify the FIM, specifically, we partly summarize parameters of a convolutional kernel by taking the mean value and ignore bias parameters as well as some of the early network layers. Furthermore, we disregard the parameters of the final layer, as their shapes may differ from task to task. In contrast to Achille et al.~\cite{achille2019task2vec}, we calculate the FIM directly and chose a different training setup based on the data augmentation and sampling strategy detailed in Sec. \ref{sec_data}. Generally speaking, the FIM is a measure for how much information the predictions of a model carry about its parameters: If a parameter of the probe-network is highly decisive for the predictions made by it, then the corresponding entry in $\mathbf{e}$ will be large. To compare any two tasks $\mathfrak{T}, \mathfrak{T}'$, we compute the Fisher embedding distance  $d_{\mathrm{FED}}(\mathfrak{T}, \mathfrak{T}')$ as the cosine distance of the embeddings. This approach is the only one incorporating the labels into the distance function.

\subsubsection{Sample-based embeddings} For the second kind of embedding we proceed as follows: For a given task $\mathfrak{T}$, we use a fixed pretrained model (the same configuration as in the previous method) to extract deep feature vectors $\mathbf{v} \in \mathbb{R}^{n}$ (n = 512 in our case) from sampled images of that task, where $\mathbf{v}$ represents the activations prior to the classification layer. To ensure fair comparison, we use exactly as many samples $m$ as during the classifier training for the FIM computation ($m = 10,000$). The sample-feature matrix $M$ may then be further processed to generate the embeddings.

\paragraph{Maximum Mean Discrepancy (MMD)} is the largest difference in expectations over functions in the unit ball of a reproducing kernel Hilbert space (RKHS) \cite{Gretton2012AKT}. It is calculated on a sample base, hence using the full matrix $M$, which serves itself as embedding in this case. Given two sample-based distributions $p, p'$ on $\mathbb{R}^n$, the MMD is an estimator for the maximal discrepancy between expectation operators on them (which is zero if $p = p'$) and can be computed as 
\begin{equation}
    d_{\mathrm{MMD}}(\mathfrak{T}, \mathfrak{T}') = \mathbb{E}_{x_1, x_2 \sim p} k(x_1, x_2) - 2\mathbb{E}_{x \sim p, y \sim p'}k(x, y) + \mathbb{E}_{y_1, y_2 \sim p'} k(y_1, y_2),
\end{equation}
where for $k(\cdot, \cdot)$ we choose the so called Cauchy kernel $k(x, y) = ( 1+ \|x - y\|^2\sigma^{-2})^{-1}$ with hyperparameter $\sigma$.

\paragraph{Kullback-Leibler Divergence (KLD)} is a concept closely connected to Fisher Information \cite{Dabak2002RelationsBK}. To leverage the KL-divergence for sample-based comparison, we base our work on an approach by Bhattacharjee et al. \cite{Bhattacharjee2020P2LPT}: Computing the arithmetic mean across all samples on $M$ and normalizing the outcome generates a discrete probability distribution $p$ over the finite space $\{1, ..., n\}$, which serves as a low dimensional embedding. Applying the entropy-based KL-divergence upon two of these distributions $p, p'$ defines the KL-distance
\begin{equation}
    d_{\mathrm{KLD}}(\mathfrak{T}, \mathfrak{T}') := \mathrm{KL}(p, p')= \sum_{1 \leq j \leq n} p(j) \log \frac{p(j)}{p'(j)}
\end{equation}
of their corresponding tasks $\mathfrak{T}$ and $\mathfrak{T}'$. The KL-distance is the only asymmetric distance function we investigate.

\paragraph{Earth Mover's Distance (EMD)} is another approach for measuring distances between distributions and originates from the theory of optimal transport \cite{Rubner1998AMF}. It also uses the full matrix $M$ as embedding. The rationale behind this approach is to quantify the minimal energy necessary to move the mass of one distribution to form another. More formally, given two distributions $p$ and $p'$ on $\mathbb{R}$, the first Wasserstein distance is defined as 
\begin{equation}
    W(p, p'):= \inf_{\gamma \in \Gamma(p, p')} \mathbb{E}_{(x, y) \sim \gamma} |x - y|,
\end{equation}
where $\Gamma(p, p')$ represents the set of distributions whose marginals are $p$ and $p'$ on the first and second factors respectively. Since the high-dimensional computation of $W$ is complex, we restrict ourselves to the $1$D case and average these distances across all feature dimensions of $M$.\footnote{We use the SciPy \cite{2020SciPy-NMeth} implementation of the Wasserstein distance.} More precisely, if $p_j, p'_j$ represent the distributions of $j$-th feature dimension ($1 \leq j \leq n$) in the samples from tasks $\mathfrak{T}, \mathfrak{T}'$, then 
\begin{equation}
    d_{\mathrm{EMD}}(\mathfrak{T}, \mathfrak{T}') := \frac{1}{n}\sum_{1 \leq j \leq n} W(p_j, p'_j).
\end{equation}

According to our experience, the most important hyperparameters are the optimizer (e.g. learning rate) and selection of FIM elements to form the embedding for FED and kernel choice as well as kernel parameter $\sigma$ for MMD. KLD and EMD are solely dependent on the sampling strategy and feature extractor, which also holds true for FED and MMD. While we did not perform a systematic grid search to optimize all of them in parallel, we individually refined them in the course of method development using tasks not in the test set of tasks applied in this work.

\subsection{Data}\label{sec_data} 

We assessed the performance of our method in the specific context of 2D medical image classification with a focus on SDS data. As we assume that medical image processing tasks can potentially benefit from data of other domains \cite{Cheplygina2018CatsOC}, we included both medical and non-medical data in this study, as summarized in  Tab. \ref{tab_datasets} within the Suppl. Materials. Overall, we used a total of 26 publicly available tasks (including 18 from biomedical imaging) with highly varying sample sizes (ranging from 434 to 184,498) and number of classes (ranging from 2 to 256). To obtain semantic meta labels for the tasks, we assigned them to their respective \textit{domains} representing the application context. The medical tasks were further assigned an additional \textit{task type} meta label based on the target structure that the image processing task focuses on (e.g. medical instrument, pathology, artefact). The labels are also given in Tab. \ref{tab_datasets}.
All data sets were split into training and validation set, either according to their standard splits, or (if not available) according to a random 80:20 split per class. We used the same data preprocessing (resizing to $224 \times 224$ pixels), data augmentation (random rotation and flipping, except for the digit classification tasks) and sampling strategies (class-balanced sampling of 10,000 images per epoch) across all tasks.

 To avoid the danger of 'model selection' based on test data, we left approx. 45\% (n = 8) of the medical imaging tasks untouched until we fully completed the processing pipeline, including the fixation of all hyperparameters. % (see previous paragraph). 
 The quantitative validation was then performed exclusively on this untouched test set of tasks.

\subsection{Experimental design}\label{sec_design}

All of the following experiments were performed five times and averaged results are reported.

\subsubsection{H1 Semantic relationships} To investigate whether task fingerprinting captures semantic relationships between tasks, we converted all 26 tasks into the \textit{fingerprint space} and analyzed whether our method results in a grouping of tasks corresponding to (1) similar domains (e.g. laparoscopic images \textit{vs.} dermatological images \textit{vs.} natural images) and (2) similar tasks (e.g. artefact detection \textit{vs.} medical instrument classification). Specifically, we paired all test tasks with every other task and computed their distance with the four approaches described in Sec. \ref{sec_methods}. We then determined intra- and inter-domain similarity based on the provided meta labels. 

\subsubsection{H2 Benefit for pretraining} The purpose of the second experiment was to assess whether the quantification of task similarity can be leveraged to select appropriate pretraining tasks. 
To this end, we proceeded as follows for each of the 8 test tasks: We sampled a (class-balanced) random subset of size 300 (train: 240, validation: 60) from the training split to simulate a data scarcity scenario. Given any of the other data sets as source task, we used a pretrained ResNet50 \cite{He2016DeepRL} model and tuned the model for 15 epochs (each $10,000$ samples) on the source task. After replacing the final classification layer, we tuned for 15 more epochs on the target task. Based on the best validation performance we evaluated the resulting model on the unseen original validation split of the target task. As a baseline, we also omitted the first step and directly tuned on the target task. Based on these results we defined the \textit{relative transfer learning (TL) success} as the tuned cross-entropy loss divided by the baseline cross-entropy loss. This makes the results for different target tasks of varying complexity and number of classes comparable.

\subsubsection{H3 Benefit for architecture selection} The purpose of the third experiment was to assess whether task fingerprints can be leveraged to select an appropriate model for a new given task. To this end, we applied the following strategy. Initially, we trained a set of $n=11$ different neural network architectures (inclusion criteria: ImageNet performance, parameter count, variety of models and implementation availability)\footnote{We used the implementations from \cite{rw2019timm} and the models
CSPNet (cspdarknet53, cspresnext50), 
ECA-Net (ecaresnet50d), 
EfficientNet (tf\_efficientnet\_b2\_ns),
DPN (dpn68b), 
MixNet (mixnet\_xl),
RegNetY (regnety\_032),
ResNeXt (swsl\_resnext50\_32x4d),
ReXNet (rexnet\_200),
VovNet2 (ese\_vovnet39b) and
Xception (xception). Please refer to the documentation of \cite{rw2019timm} for full references of these.} for all of our 18 medical imaging tasks. Training was done for 50 epochs (each $10,000$ samples again) with Adam \cite{Kingma2015AdamAM}. For each task, we then ranked the different models according to their performance, resulting in an $n$-dimensional \textit{performance trace} for each task. To quantify the quality of model transfer for a given pair of two tasks, we computed the Spearman's rank correlation coefficient on the performance traces and compared the result to the distance between the tasks in fingerprint space.

\begin{figure}[h]
  \centering
  \subfloat[Colour coding of domain]{\includegraphics[width=0.4\linewidth]{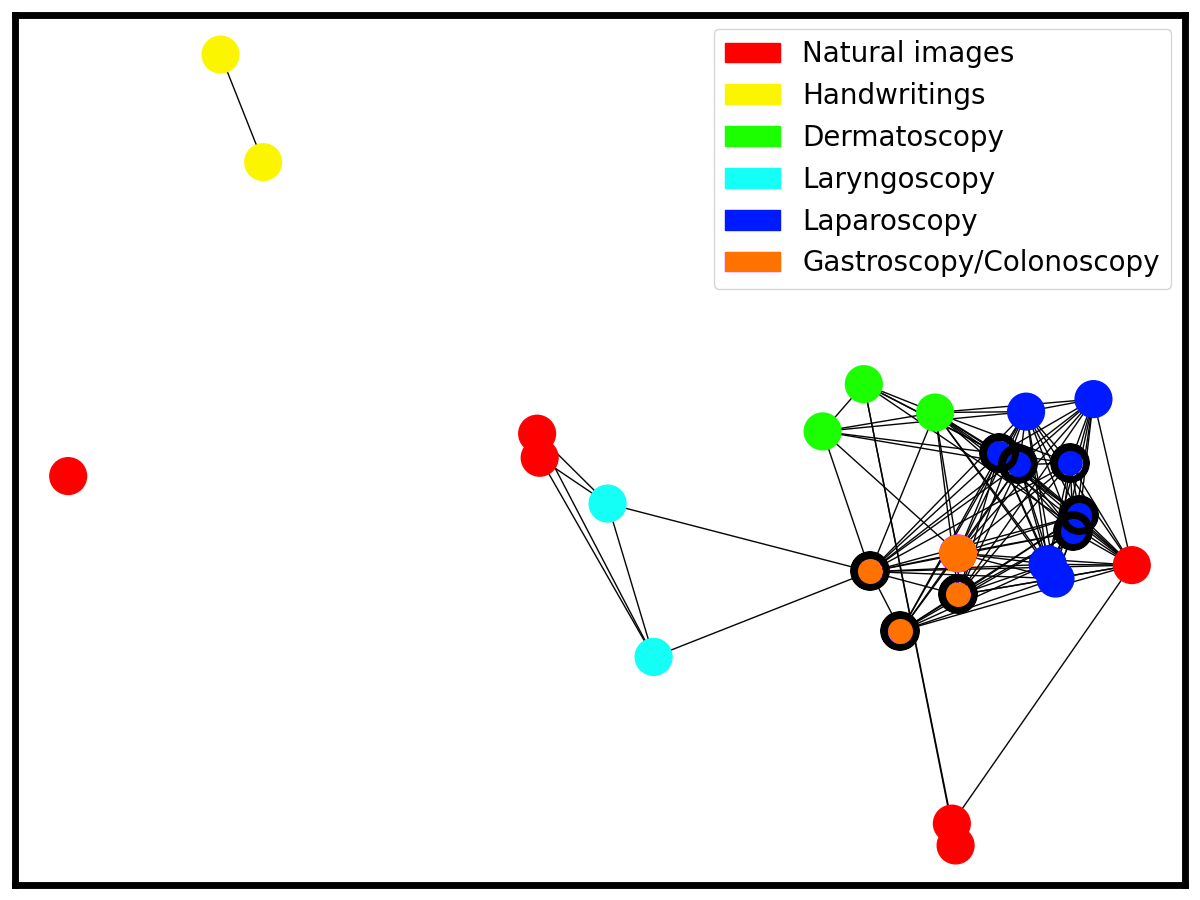}}
  \subfloat[Colour coding of task type]{\includegraphics[width=0.4\linewidth]{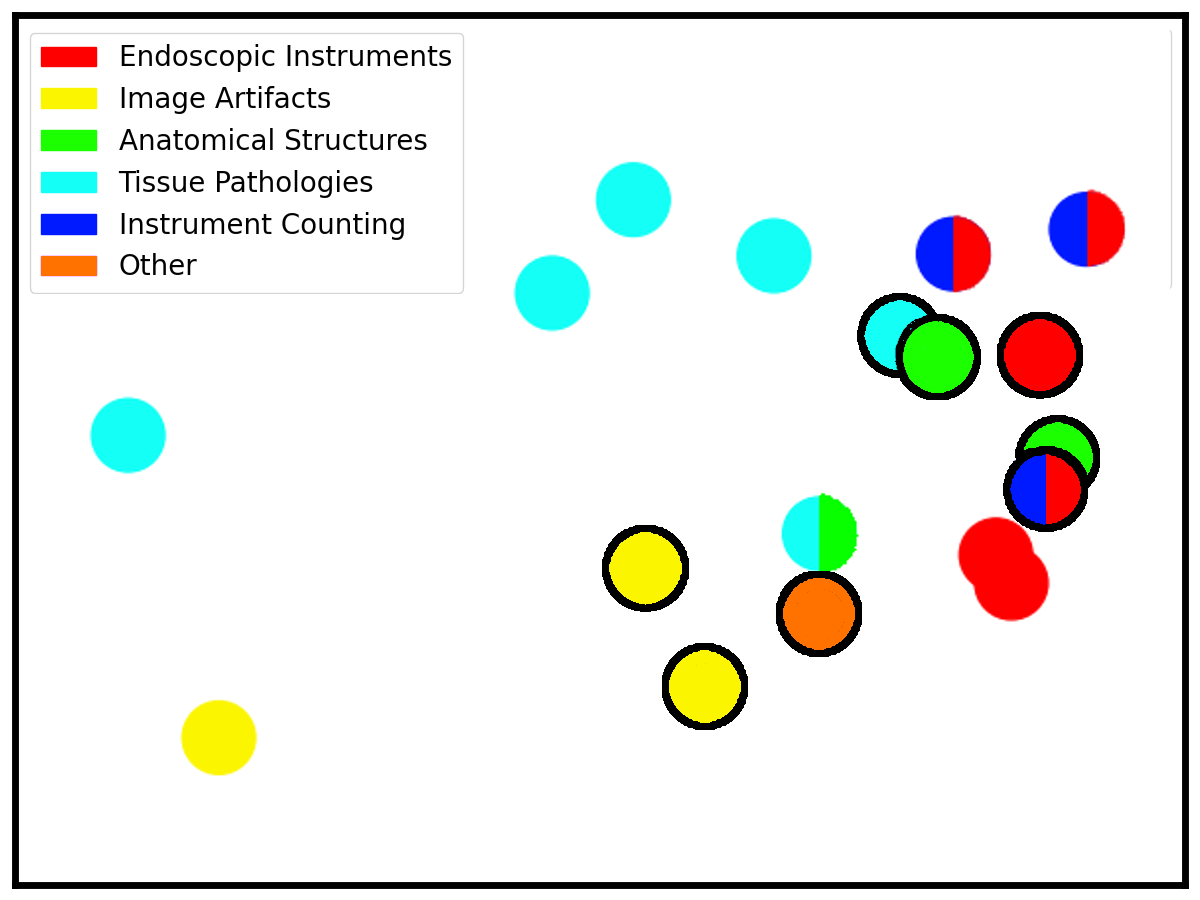}}
  \caption{2D visualizations of task fingerprints computed with the FED method. Tasks (dots) are coloured by domain ((a); all 26 tasks) and task type ((b); only medical tasks). Encircled dots represent tasks of the untouched 'test set' of tasks. Distances below a threshold of 65\% mean distance are represented by edges. Dimension reduction was achieved with the Kamada-Kawai algorithm \cite{kamada1989}. }
  \label{fig_graphs}
\end{figure}
\begin{figure}[h]
  \centering
  \includegraphics[width=0.6\linewidth]{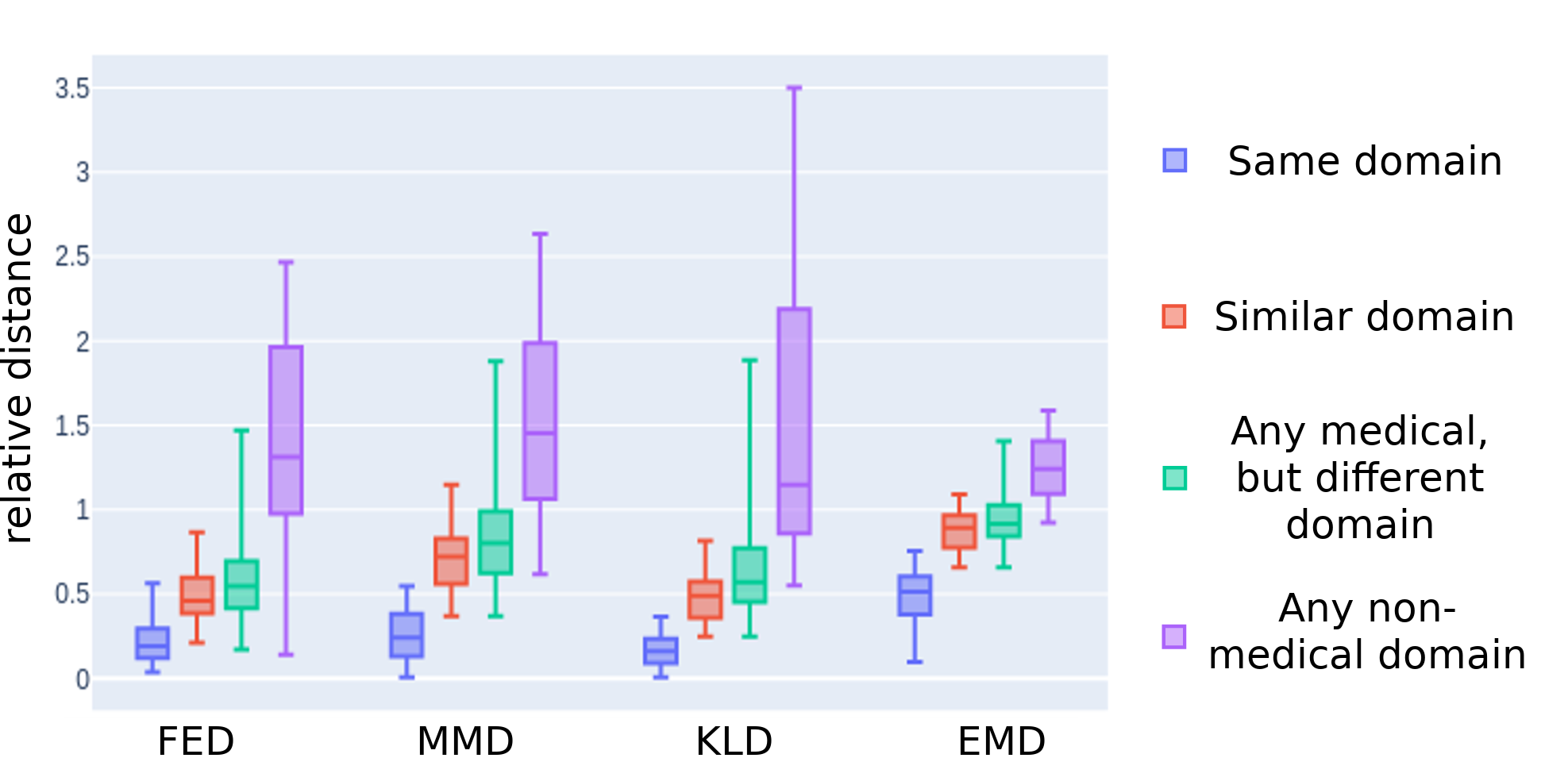}
  \caption{Task fingerprinting captures semantic similarities. The distance between tasks in fingerprint space increases with the semantic dissimilarity of domains for all four methods.}
  \label{fig_domain_distance}
\end{figure}

\section{Results}\label{sec_results}
According to our results, both the label-based method (FED) and all three sample-based methods are well-suited for capturing semantic relationships between tasks, thus confirming our hypothesis H1. 
Fig. \ref{fig_graphs} shows a 2D visualization of the task fingerprint space for the FED method, demonstrating a meaningful structuring with respect to both domain and target structure.
For example, the dermatoscopic, laparoscopic and gastroscopic tasks are clustered and related task pairs, such as (CIFAR-10, CIFAR-100), (MNIST, EMNIST) and (Caltech101, Caltech256) are close to each other in Fig. \ref{fig_graphs}a. Similarly, an analysis of the medical tasks reveals that similar tasks according to algorithm targets (e.g. pathology \textit{vs.} artefact) cluster as well (Fig. \ref{fig_graphs}b). 
 The rather qualitative results were confirmed by our quantitative assessment which shows that the distance between tasks in fingerprint space increases with the semantic dissimilarity of domains for all four methods (Fig.~\ref{fig_domain_distance}). Similarly, tasks with similar targets feature a comparatively low task distance with the FED and KLD methods, as shown in Tab. \ref{tab_target_method} in the Suppl. Materials.

\begin{figure}[h]
  \centering
  \subfloat[Benefit for pretraining]{\includegraphics[width=0.37\linewidth]{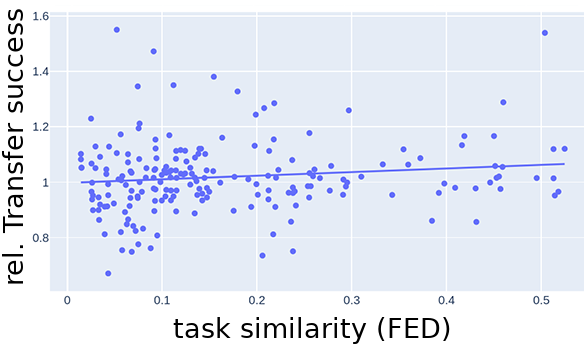}}
  \subfloat[Benefit for model selection]{\includegraphics[width=0.4\linewidth]{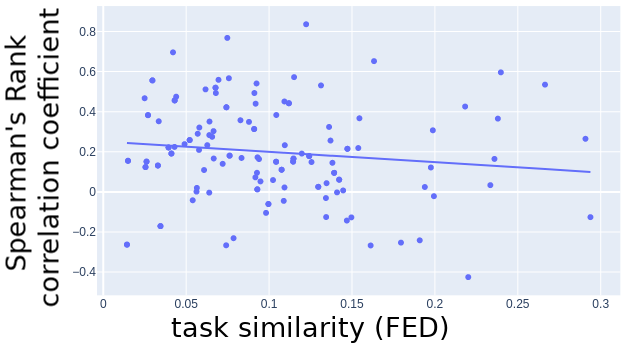}}
  \caption{Task fingerprinting benefits knowledge transfer for both pretraining and model selection.  \textbf{(a)} Relative transfer learning success and \textbf{(b)} Spearman's rank correlation coefficient (see Sec. \ref{sec_results}) both plotted against FED distance. Trendline has been created using ordinary least squares regression.}
  \label{fig_fed_eval}
\end{figure}

\begin{figure}[!h]
  \centering
  \subfloat[MMD]{\includegraphics[width=0.3\linewidth]{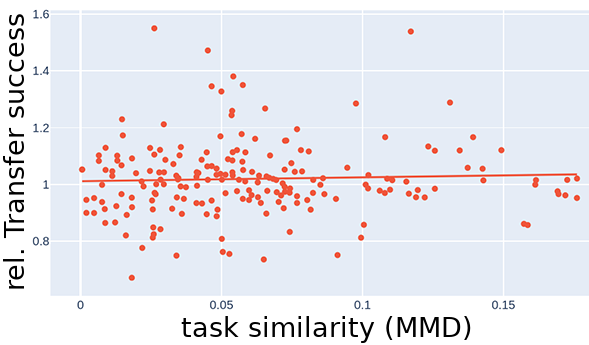}}
  \subfloat[KLD]{\includegraphics[width=0.3\linewidth]{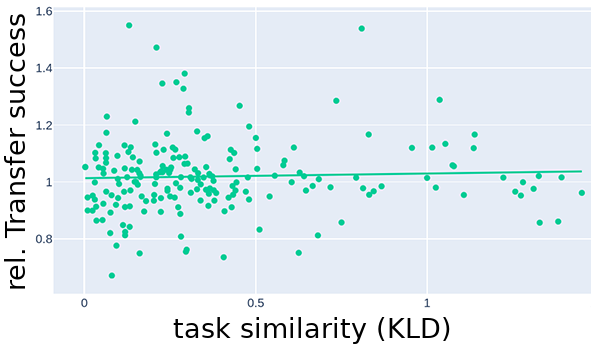}}
  \subfloat[EMD]{\includegraphics[width=0.3\linewidth]{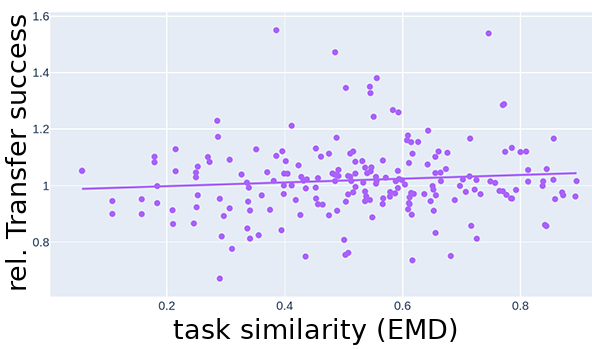}}\
  \caption{Benefit of fingerprinting for pretraining for the sample-based methods (complementing Fig.~\ref{fig_fed_eval}a in main text). In contrast to the FED method, the sample-based methods do not yield clear insights on positive transfer. Trend lines were generated using ordinary least squares regression.}
  \label{fig_sup_tl}
\end{figure}

\begin{figure}[!h]
  \centering
  \subfloat[MMD]{\includegraphics[width=0.3\linewidth]{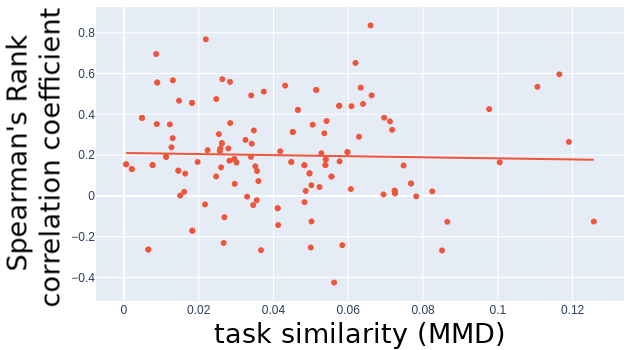}}
  \subfloat[KLD]{\includegraphics[width=0.3\linewidth]{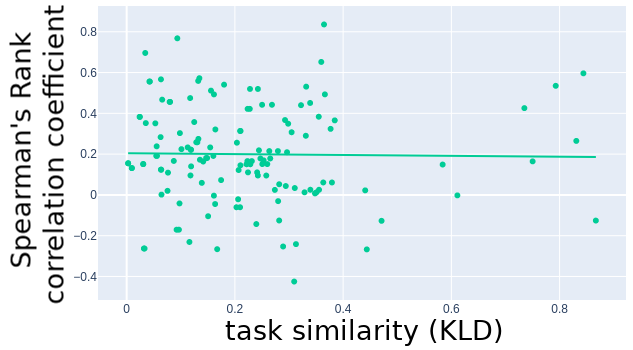}}
  \subfloat[EMD]{\includegraphics[width=0.3\linewidth]{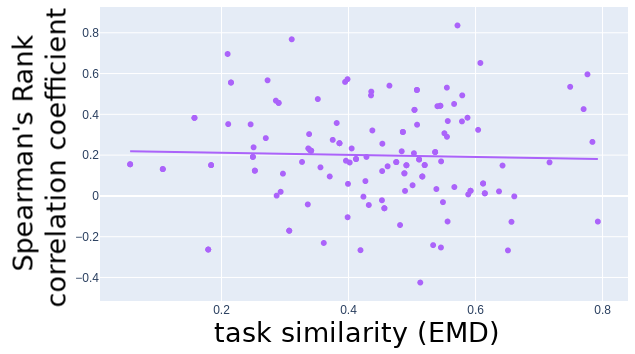}}
  \caption{Effect of fingerprinting for model selection for the sample-based methods (complementing Fig. \ref{fig_fed_eval}b in main text). In contrast to the FED method, the sample-based methods do not yield clear insights on model transferability. Trend lines were generated using ordinary least squares regression.}
  \label{fig_sup_model}
\end{figure}

A further important result of our study is that task fingerprinting with the FED method can be used for selecting appropriate tasks for pretraining or model selection, thus confirming hypotheses H2 and H3 (Fig. \ref{fig_fed_eval}). To compare the four methods with respect to these aspects, we computed the Spearman's Rank correlation coefficients $\rho$ and performed significance tests for all of them. The three sample-based methods perform similarly for the pretraining task (H2) with $\rho_{\mathrm{MMD}} = 0.04\ (p=0.53)$, $\rho_{\mathrm{KLD}} = 0.06\ (p=0.36)$ and $\rho_{\mathrm{EMD}} = 0.07\ (p=0.27)$ (see Suppl. Materials Fig. \ref{fig_sup_tl}) as well as for the performance rankings of neural network architectures (H3) with $\rho_{\mathrm{MMD}} = -0.06\ (p=0.45)$, $\rho_{\mathrm{KLD}} = -0.07\ (p=0.36)$ and $\rho_{\mathrm{EMD}} = -0.06\ (p=0.42)$ (see Suppl. Materials Fig. \ref{fig_sup_model}). In contrast, the FED method outperforms the other methods in both experiments with $\rho_{\mathrm{FED}} = 0.12\ (p=0.07)$ for pretraining and $\rho_{\mathrm{FED}} = -0.16\ (p=0.05)$ for model selection. 

\section{Discussion} 
To our knowledge, this paper is the first to address the challenge of task similarity quantification in the field of SDS. According to our results, our approach is well-suited for capturing semantic relationships between tasks and can be leveraged for selecting appropriate tasks for pretraining and model selection. Incorporation of labels in the fingerprints turned out to be particularly useful. 

The closest work to ours in the field of biomedical image analysis was presented by Cheplygina et al. \cite{Cheplygina2017ExploringTS} (see Sec. \ref{sec_intro}). We built upon this general idea by exploring approaches that produce fingerprints invariant to the (other) available data sets and that are thus inherently scalable. 
Outside of the field of biomedical imaging, task similarity estimation is an increasingly active field of research, and our research has been inspired by previous work in the general machine learning community~\cite{achille2019task2vec,Bhattacharjee2020P2LPT}. However, the research is still in its infancy, and no widely accepted approaches have been established.

It should be noted that there are potentially important parameters for knowledge transfer that we did not incorporate into our methods (e.g. number of samples) and that no extensive ablation studies for hyperparameter optimization were performed. The rationale for this strategy was that we are seeking a way to quantify the relationship between tasks rather than optimizing a model for a given task. While our initial work is encouraging, further work is required (1) to enable drawing broad conclusions, (2) to transfer the methodology to other types of tasks (e.g. semantic segmentation) and (3) to implement the approach in a complete meta learning framework. 

In the long-term, the ability to quantify similarities between tasks could evolve as an enabling technique to leverage the full potential of meta learning for the field of biomedical imaging, and thus pave the way for translating biomedical image analysis research into clinical practice.

\subsubsection{Acknowledgements}
We would like to thank our colleagues Minu Dietlinde Tizabi, Tim Adler, Thuy Nuong Tran, Tobias Ross and Lucas-Raphael Müller for their valuable feedback on the drafts of this work. This project has been funded by the Surgical Oncology Program of the National Center for Tumor Diseases (NCT) Heidelberg. The present contribution is also supported by the Helmholtz Imaging Platform (HIP), a platform of the Helmholtz Incubator on Information and Data Science.
\bibliographystyle{splncs04}
\bibliography{references_tiny.bib}

\begin{thebibliography}{10}
\providecommand{\url}[1]{\texttt{#1}}
\providecommand{\urlprefix}{URL }
\providecommand{\doi}[1]{https://doi.org/#1}

\bibitem{achille2019task2vec}
{Achille}, A., {Lam}, M., et~al.: {Task2Vec: Task Embedding for Meta-Learning}.
  ArXiv  (2019)

\bibitem{achille2019information}
{Achille}, A., {Paolini}, G., et~al.: {The Information Complexity of Learning
  Tasks, their Structure and their Distance}. ArXiv  (2019)

\bibitem{Ali2019EndoscopyAD}
Ali, S., Zhou, F., et~al.: Endoscopy artifact detection (ead 2019) challenge
  dataset. ArXiv  (2019)

\bibitem{Allan20192017RI}
Allan, M., Shvets, A., et~al.: 2017 robotic instrument segmentation challenge.
  ArXiv  (2019)

\bibitem{Bhattacharjee2020P2LPT}
Bhattacharjee, B., Codella, N.C.F., et~al.: P2l: Predicting transfer learning
  for images and semantic relations. CVPR Workshops  (2020)

\bibitem{Borgli2020HyperKvasirAC}
Borgli, H., Thambawita, V., et~al.: Hyperkvasir, a comprehensive multi-class
  image and video dataset for gastrointestinal endoscopy. Scientific Data
  (2020)

\bibitem{Cheplygina2018CatsOC}
Cheplygina, V.: Cats or cat scans: transfer learning from natural or medical
  image source datasets? ArXiv  (2018)

\bibitem{Cheplygina2019NotsosupervisedAS}
Cheplygina, V., Bruijne, M., et~al.: Not‐so‐supervised: A survey of
  semi‐supervised, multi‐instance, and transfer learning in medical image
  analysis. Medical Image Analysis  (2019)

\bibitem{Cheplygina2017ExploringTS}
Cheplygina, V., Moeskops, P., et~al.: Exploring the similarity of medical
  imaging classification problems. In: CVII-STENT/LABELS@MICCAI (2017)

\bibitem{Cohen2017EMNISTAE}
Cohen, G., Afshar, S., et~al.: Emnist: an extension of mnist to handwritten
  letters. ArXiv  (2017)

\bibitem{Combalia2019BCN20000DL}
Combalia, M., Codella, N.C.F., et~al.: Bcn20000: Dermoscopic lesions in the
  wild. ArXiv  (2019)

\bibitem{Dabak2002RelationsBK}
Dabak, A., Johnson, D.: Relations between kullback-leibler distance and fisher
  information (2002)

\bibitem{Deng2009ImageNetAL}
Deng, J., Dong, W., et~al.: Imagenet: A large-scale hierarchical image
  database. 2009 IEEE Conference on Computer Vision and Pattern Recognition
  (2009)

\bibitem{Faria2019LightFI}
Faria, S.M., Filipe, J.N., et~al.: Light field image dataset of skin lesions.
  2019 41st Annual International Conference of the IEEE Engineering in Medicine
  and Biology Society (EMBC)  (2019)

\bibitem{FeiFei2004LearningGV}
Fei-Fei, L., Fergus, R., et~al.: Learning generative visual models from few
  training examples: An incremental bayesian approach tested on 101 object
  categories. In: CVPR Workshops (2004)

\bibitem{Gretton2012AKT}
Gretton, A., Borgwardt, K., et~al.: A kernel two-sample test. J. Mach. Learn.
  Res.  (2012)

\bibitem{Griffin2007Caltech256OC}
Griffin, G., Holub, A., et~al.: Caltech-256 object category dataset (2007)

\bibitem{Gutman2018SkinLA}
Gutman, D., Codella, N.C.F., et~al.: Skin lesion analysis toward melanoma
  detection: A challenge at the 2017 international symposium on biomedical
  imaging (isbi), hosted by the international skin imaging collaboration
  (isic). 2018 IEEE 15th International Symposium on Biomedical Imaging (ISBI
  2018)  (2018)

\bibitem{He2016DeepRL}
He, K., Zhang, X., et~al.: Deep residual learning for image recognition. 2016
  IEEE Conference on Computer Vision and Pattern Recognition (CVPR)  (2016)

\bibitem{Hospedales2020MetaLearningIN}
Hospedales, T.M., Antoniou, A., et~al.: Meta-learning in neural networks: A
  survey. ArXiv  (2020)

\bibitem{kamada1989}
Kamada, T., Kawai, S.: An algorithm for drawing general undirected graphs. Inf.
  Process. Lett.  (1989)

\bibitem{skin-lesion}
{Kawahara}, J., {Daneshvar}, S., et~al.: Seven-point checklist and skin lesion
  classification using multitask multimodal neural nets. IEEE Journal of
  Biomedical and Health Informatics  (2019)

\bibitem{KhoslaYaoJayadevaprakashFeiFei_FGVC2011}
Khosla, A., Jayadevaprakash, N., et~al.: Novel dataset for fine-grained image
  categorization. In: First Workshop on Fine-Grained Visual Categorization,
  IEEE Conference on Computer Vision and Pattern Recognition. Colorado Springs,
  CO (2011)

\bibitem{Kingma2015AdamAM}
Kingma, D.P., Ba, J.: Adam: A method for stochastic optimization. CoRR  (2015)

\bibitem{Krizhevsky2009LearningML}
Krizhevsky, A.: Learning multiple layers of features from tiny images (2009)

\bibitem{LeCun1998GradientbasedLA}
LeCun, Y., Bottou, L., et~al.: Gradient-based learning applied to document
  recognition (1998)

\bibitem{Leibetseder2020GLENDAGL}
Leibetseder, A., Kletz, S., et~al.: Glenda: Gynecologic laparoscopy
  endometriosis dataset. In: MMM (2020)

\bibitem{Leibetseder2018Lapgyn4AD}
Leibetseder, A., Petscharnig, S., et~al.: Lapgyn4: a dataset for 4 automatic
  content analysis problems in the domain of laparoscopic gynecology.
  Proceedings of the 9th ACM Multimedia Systems Conference  (2018)

\bibitem{MaierHein2014CanMO}
Maier-Hein, L., Mersmann, S., et~al.: Can masses of non-experts train highly
  accurate image classifiers? - a crowdsourcing approach to instrument
  segmentation in laparoscopic images. MICCAI International Conference on
  Medical Image Computing and Computer-Assisted Intervention  (2014)

\bibitem{maier-hein_sds_2020}
Maier-Hein, L., Eisenmann, M., et~al.: Surgical {Data} {Science} — from
  {Concepts} to {Clinical} {Translation}. ArXiv  (Oct 2020)

\bibitem{Moccia2017ConfidentTL}
Moccia, S., Momi, E.D., et~al.: Confident texture-based laryngeal tissue
  classification for early stage diagnosis support. Journal of Medical Imaging
  (2017)

\bibitem{Moccia2018LearningbasedCO}
Moccia, S., Vanone, G.O., et~al.: Learning-based classification of informative
  laryngoscopic frames. Computer methods and programs in biomedicine  (2018)

\bibitem{Netzer2011ReadingDI}
Netzer, Y., Wang, T., et~al.: Reading digits in natural images with
  unsupervised feature learning (2011)

\bibitem{Pogorelov:2017:NBP:3083187.3083216}
Pogorelov, K., Randel, K.R., et~al.: Nerthus: A bowel preparation quality video
  dataset. In: Proceedings of the 8th ACM on Multimedia Systems Conference.
  MMSys'17, ACM, New York, NY, USA (2017)

\bibitem{Raghu2019TransfusionUT}
Raghu, M., Zhang, C., et~al.: Transfusion: Understanding transfer learning for
  medical imaging. In: NeurIPS (2019)

\bibitem{ross2020robust}
Ross, T., Reinke, A., et~al.: Robust medical instrument segmentation challenge
  2019 (2020)

\bibitem{Ro2018ExploitingTP}
Ro{\ss}, T., Zimmerer, D., et~al.: Exploiting the potential of unlabeled
  endoscopic video data with self-supervised learning. International Journal of
  Computer Assisted Radiology and Surgery  (2018)

\bibitem{Rubner1998AMF}
Rubner, Y., Tomasi, C., et~al.: A metric for distributions with applications to
  image databases. Sixth International Conference on Computer Vision  (1998)

\bibitem{Tschandl2018TheHD}
Tschandl, P., Rosendahl, C., et~al.: The ham10000 dataset, a large collection
  of multi-source dermatoscopic images of common pigmented skin lesions.
  Scientific Data  (2018)

\bibitem{Twinanda2017EndoNetAD}
Twinanda, A.P., Shehata, S., et~al.: Endonet: A deep architecture for
  recognition tasks on laparoscopic videos. IEEE Transactions on Medical
  Imaging  (2017)

\bibitem{Vilalta2005APV}
Vilalta, R., Drissi, Y.: A perspective view and survey of meta-learning.
  Artificial Intelligence Review  (2005)

\bibitem{2020SciPy-NMeth}
Virtanen, P., Gommers, R., et~al.: {{SciPy} 1.0: Fundamental Algorithms for
  Scientific Computing in Python}. Nature Methods  (2020)

\bibitem{rw2019timm}
Wightman, R.: Pytorch image models (2019). \doi{10.5281/zenodo.4414861}

\bibitem{Yuan2020FewIE}
Yuan, P., Mobiny, A., et~al.: Few is enough: Task-augmented active
  meta-learning for brain cell classification. In: MICCAI (2020)

\bibitem{Zhang2020GeneralizingDL}
Zhang, L., Wang, X., et~al.: Generalizing deep learning for medical image
  segmentation to unseen domains via deep stacked transformation. IEEE
  Transactions on Medical Imaging  (2020)

\end{thebibliography}

\newpage
\section{Supplementary Material}
\begin{table}[h]
\caption{Average intra-cluster similarities for task types. For comparison the methods have been relativized by their mean values over all 26 tasks. Results based solely on 'test tasks' (lower is better).}\label{tab_target_method}
\begin{tabular}{lccccc}
%\hline
Target subject &  FED & MMD & KLD & EMD & test/total tasks \\
\hline
End. instruments &  $0.21 \pm 0.12$ & $0.17 \pm 0.14$ & $\mathbf{0.10 \pm 0.08}$ & $0.39 \pm 0.16$ & 2 / 6\\
Counting instr. & $0.35 \pm 0.08$ & $0.39 \pm 0.00$ & $\mathbf{0.24 \pm 0.00}$ & $0.64 \pm 0.00$ & 1 / 3\\
Tissue pathology & $\mathbf{0.54 \pm 0.33}$ & $0.95 \pm 0.32$ & $0.89 \pm 0.39$ & $0.98 \pm 0.16$ & 1 /  6\\
Anatomical struct. & $\mathbf{0.30 \pm 0.11}$ & $0.52 \pm 0.25$ & $0.35 \pm 0.16$ & $0.72 \pm 0.18$ & 2 /  3\\
Image artifacts & $0.33 \pm 0.20$ & $0.27 \pm 0.19$ & $\mathbf{0.21 \pm 0.15}$ & $0.50 \pm 0.29$ & 2 / 3\\
\hline
Mean & $\mathbf{0.35 \pm 0.17}$ & $0.46 \pm 0.18$ & $0.36 \pm 0.15$ & $0.65 \pm 0.16$ & - \\
%\hline
\end{tabular}
\end{table}

\begin{table}[h]
  \caption{Overview on data sets used ordered by size. Tasks written \textbf{bold} belong to our 'test set'. Domain-Labels: \textbf{Der}: Dermatoscopy, \textbf{Gas}: Gastroscopy and Colonoscopy, \textbf{Han}: Handwritings, \textbf{Lap}: Laparoscopy, \textbf{Lar}: Laryngoscopy, \textbf{Nat}: Natural Images. Task type labels: \textbf{Art}: Image artifacts, \textbf{Ana}: Anatomical structures, \textbf{Cou}: Counting surgical instruments, \textbf{Ins}: Endoscopic instruments, \textbf{Tis}: Tissue pathology.}\label{tab_datasets}
  \centering
  \begin{tabular}{lclcc}
    \toprule
    % \multicolumn{2}{c}{Part}                  \\
    % \cmidrule(r){1-2}
    Name \& Reference     & Size  & Labels  & \# Classes & Note\\
    \midrule
    SKINL2 \cite{Faria2019LightFI}     & 434  & Der, Tis  & 6 & -\\
    NBI-InfFrames \cite{Moccia2018LearningbasedCO} & 720 & Lar, Art & 4 & - \\
    Skin-Lesion \cite{skin-lesion} & 1011 & Der, Tis & 9 & \begin{tabular}[x]{@{}c@{}}we merged some\\ similar classes\end{tabular}\\
    Laryngeal \cite{Moccia2017ConfidentTL} & 1320 & Lar, Tis & 4 & - \\
    EndoVis17-RIS \cite{Allan20192017RI}     & 1800  & Lap, Ins, Cou  & 4 & \begin{tabular}[x]{@{}c@{}}used as \\counting task\end{tabular}\\
    %\begin{tabular}[x]{@{}c@{}}binary detection task,\\only Vessel Sealer\end{tabular}\\
    \textbf{EAD19} \cite{Ali2019EndoscopyAD} & 2147 & Gas, Art & 2 & \begin{tabular}[x]{@{}c@{}} we sep. used Blur \& \\ Specularity presence\end{tabular}\\
    \begin{tabular}[x]{@{}l@{}}\textbf{LapGyn4} \cite{Leibetseder2018Lapgyn4AD} -\\\textbf{Anatomical Structures}\end{tabular} & 2728 & Lap, Ana & 5 & -\\
    \begin{tabular}[x]{@{}l@{}}\textbf{LapGyn4} \cite{Leibetseder2018Lapgyn4AD} -\\\textbf{Actions on Anatomy}\end{tabular} & 4782 & Lap, Ana & 4 & -\\
    \textbf{NERTHUS} \cite{Pogorelov:2017:NBP:3083187.3083216} & 5525 & Gas & 4 & -\\
    Caltech101 \cite{FeiFei2004LearningGV}     & 8677  & Nat & 101 & -\\
    RobustMIS \cite{ross2020robust} & 8863 & Lap, Ins, Cou & 2 & \begin{tabular}[x]{@{}c@{}}counting task,\\ using Stage 3\end{tabular}\\
    Hyperkvasir \cite{Borgli2020HyperKvasirAC} & 10,215 & Gas, Ana, Tis & 14 & \begin{tabular}[x]{@{}c@{}}excluded 9 \\small classes\end{tabular}     \\
    Stanford Dogs \cite{KhoslaYaoJayadevaprakashFeiFei_FGVC2011}     & 20,580  & Nat  & 120 & -\\
    \begin{tabular}[x]{@{}l@{}}\textbf{LapGyn4} \cite{Leibetseder2018Lapgyn4AD,Twinanda2017EndoNetAD} -\\\textbf{Instrument Count}\end{tabular} & 21,424 & Lap, Ins, Cou & 4 & -\\
    ISIC19 \cite{Tschandl2018TheHD,Gutman2018SkinLA,Combalia2019BCN20000DL} & 25,331  & Der, Tis  & 8 & - \\
    \textbf{GLENDA} \cite{Leibetseder2020GLENDAGL} & 25,408 & Lap, Tis & 5 & \begin{tabular}[x]{@{}c@{}}omitted multi-class\\ frames\end{tabular}\\
    Caltech256 \cite{Griffin2007Caltech256OC}     & 29,780  & Nat & 256 & -\\
    \begin{tabular}[x]{@{}l@{}}\textbf{LapGyn4} \cite{Leibetseder2018Lapgyn4AD} -\\\textbf{Surgical Actions}\end{tabular}& 30,682 & Lap, Ins & 8 & -\\
    CIFAR-10 \cite{Krizhevsky2009LearningML}     & 60,000  & Nat  & 10 & -\\
    CIFAR-100  \cite{Krizhevsky2009LearningML}     & 60,000  & Nat  & 100 & -\\
    MNIST \cite{LeCun1998GradientbasedLA}     & 70,000  & Han & 10 & -\\
    SVHN \cite{Netzer2011ReadingDI} & 99,289 & Nat & 10 & -\\
    EMNIST \cite{Cohen2017EMNISTAE}     & 131,600  & Han & 47 & -\\
    Cholec80 \cite{Twinanda2017EndoNetAD} & 184,498 & Lap, Ins & 2 & \begin{tabular}[x]{@{}c@{}}we sep. used Hook \& \\Grasper tool presence\end{tabular}\\
    \bottomrule
  \end{tabular}
  \label{tab_data sets}
\end{table}

\end{document}